\def\year{2018}\relax
\documentclass[letterpaper]{article} 
\usepackage{aaai18}  
\usepackage{times}  
\usepackage{helvet}  
\usepackage{courier}  
\usepackage{url}  
\usepackage{graphicx}  
\usepackage{subfigure}
\usepackage[x11names,dvipsnames,table]{xcolor} 
\usepackage{colortbl}

\newcommand{\word}[1]{\texttt{#1}}

\newcommand{\y}{{\bf y}}
\newcommand{\x}{{\bf x}}
\newcommand{\bc}{{\bf c}}
\newcommand{\z}{{\bf z}}

\begin{document}
%
\title{A Novel Document Generation Process for Topic Detection \\ based on Hierarchical Latent Tree Models}
\author{Peixian Chen, ~
Zhourong Chen,~
Nevin L. Zhang \footnote{Corresponding Author.}\\
The Hong Kong University of Science and Technology \{pchenac,zchenbb,lzhang@cse.ust.hk\}\\}
\maketitle
\begin{abstract}

We propose a novel document generation process based on hierarchical latent tree models (HLTMs) learned from data. An HLTM has a layer of observed word variables at the bottom and multiple layers of  latent variables on top. For each document, we first sample values for the latent variables layer by layer via logic sampling, then
draw relative frequencies for the words conditioned on the values of the latent variables, and finally generate words for the document using the relative word frequencies. The motivation for the work is to  take word counts into consideration with HLTMs. In comparison with LDA-based hierarchical document generation processes, the new process achieves drastically  better model fit with much fewer parameters. It also yields more meaningful topics and topic hierarchies.
It is the new state-of-the-art for the hierarchical topic detection.

\end{abstract}

\section{Introduction}
\begin{figure*}[t]
\begin{center}
\includegraphics[width=12cm]{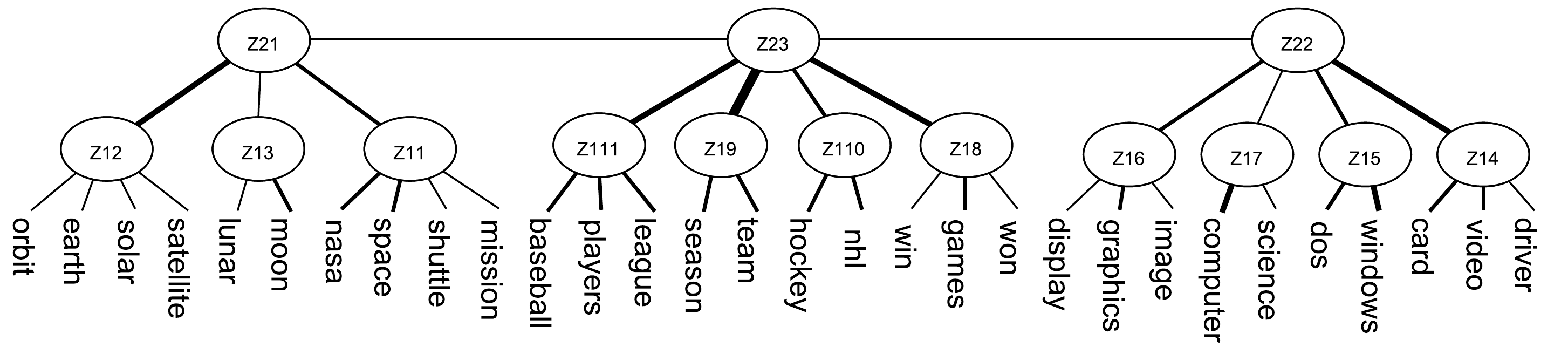}
\caption{An example hierarchical latent tree model from ~\cite{chen2016progressive}, learned by HLTA on a toy text dataset.}
\label{fig.hltm}
\end{center}
\end{figure*}

The objective of {\em hierarchical topic detection} is, given a corpus of
documents, to obtain a tree of topics with more general topics at high levels of the tree
and more specific topics at low levels of the tree.
Several hierarchical topic detection methods have been proposed based  on latent Dirichlet  allocation (LDA)~\cite{blei2003latent}, including the hierarchical latent Dirichlet allocation (hLDA) model ~\cite{blei10nested}, the hierarchical Pachinko allocation model (hPAM) ~\cite{li2006pachinko,mimno2007mixtures}, and  the nested hierarchical Dirichlet process (nHDP)~\cite{paisley2015nested} .

A very different method named hierarchical latent tree analysis (HLTA) is recently proposed by~\cite{liu2014hierarchical,chen2016progressive,chen2017latent}. HLTA learns  models such as the one shown in Figure~\ref{fig.hltm}, where there is a layer of observed variables at the bottom, and one or more layers of latent variables on top. The variables are connected up to form a tree. The model is hence called a {\em hierarchical latent tree model (HLTM)}. The observed variables  are binary and represent the absence or presence of words in documents.  The latent variables are also binary variables. They are introduced during data analysis to explain co-occurrence patterns. For example, $z14$  explains the probabilistic co-occurrences of the words \word{card}, {\tt video} and {\tt driver}; $z16$ explains the co-occurrences of {\tt display}, {\tt graphics} and {\tt image}; and $z22$   explains the probabilistic co-occurrence of the patterns represented by $z14$, $z15$, $z16$ and $z17$.

HLTMs  is a generalization of latent class models (LCMs)\cite{bartholomew99latent}, which is a type of finite mixture models for discrete data. In a finite mixture model, there is one latent variable and it is used to partition objects into soft clusters. Similarly, in an HLTM, each latent variable  partitions all the documents into two clusters. One of the clusters consists of the documents that contain, in a probabilistic sense, the words in subtree rooted at the latent variable. It is interpreted as a topic. The other cluster is viewed as the background. In this manner each latent variable gives one topic.

Topics given by some of the latent variables in Figure~\ref{fig.hltm} are listed below. For example, $z{14}$ gives a topic that consists of 12\% of the documents, and  the words {\tt card}, {\tt video} and {\tt driver} occur with relatively high probabilities inside the topic and relatively low probabilities outside. Note that, for $z22$,  only a subset of words in its subtree are used when characterizing the topic. The reader is referred to~\cite{chen2017latent} for how the words for characterizing a topic are picked and ordered.

{\small
\begin{tabbing}
\=\hspace*{1cm} \=\hspace*{0.7cm} \=\hspace*{0.7cm} \=\hspace*{0.5cm}\= \kill
\> \>  $z{22}$ \> [0.24] {\tt windows} {\tt card} {\tt graphics} {\tt video} {\tt dos}    \\
\>\>\> $z{14}$ \> [0.12]  {\tt card} {\tt video} {\tt driver} \\
\>\>\> $z{15}$ \> [0.15]  {\tt windows} {\tt dos} \\
\>\>\> $z{16}$ \> [0.10]  {\tt graphics} {\tt display} {\tt image}
 \\ \>\>\> $Z{17}$ \> [0.09]  {\tt computer} {\tt science}
\end{tabbing}
}

 In general, latent variables at high levels of an HLTM capture ``long-range" word co-occurrence
patterns and hence give thematically more general topics, while those at low levels
capture ``short-range" word co-occurrence patterns and give thematically
more specific topics. For example, the topic given by $z22$ concerns several aspects of
computers, while its subtopics are each concerned with only one aspect of computers.
 In this sense, HLTA is a tool for hierarchical topic detection.

As will be discussed in the next section, HLTA differs fundamentally from the LDA-based methods. However, comparisons between them are still possible. The reason is that they both define distributions over documents and characterize  topics using lists of words. Empirical results reported by \cite{chen2016progressive,chen2017latent}
show that  HLTA significantly outperforms the LDA-based methods  in terms of model quality
as measured by held-out likelihood, and it finds more
meaningful topics and topic hierarchies.

It should be noted, however,  that
the aforementioned comparisons were conducted only on binary data. The reason is that  HLTA is unable to take word counts into consideration. In the experiments,  documents were represented as  binary vectors over the vocabulary for HLTA. For the LDA-based methods, they were represented as  bags of words, where duplicates removed such that no word appears more than once. The two representations are equivalent.

To amend the serious drawback, this paper extends HLTA so as to take word counts into consideration. Specifically, we propose a document generation model based on the model structure learned by HLTA from binary data, design a parameter learning algorithm for the new model,   and give an importance sampling method for model evaluation.
 The new method is named HLTA-c, where the letter ``c" stands for count data. We present empirical results to show that, on count data,  HLTA-c also significantly outperforms LDA-based methods in terms of both model quality and meaningfulness of topics and  topic hierarchies.



\section{Related work}
Detecting topics and topic hierarchies from large archives of documents has been one of the most active research areas in last decade. The most commonly used method is latent Dirichlet allocation (LDA)~\cite{blei2003latent}. LDA has been extended in various ways for additional modeling capabilities. Topic correlations are considered in~\cite{lafferty2006correlated,li2006pachinko}; topic evolution is modeled in~\cite{blei2006dynamic,wang2006topics,ahmed2007dynamic}; topic hierarchies are built in ~\cite{li2006pachinko,griffiths2004hierarchical,mimno2007mixtures}; side information is exploited in~\cite{andrzejewski2009,jagarlamudi2012};  and so on.

A fundamental difference between HLTA/HLTA-c and the LDA-based methods for hierarchical topic detection is that observed variables in HLTA/HLTA-c correspond to words in the vocabulary, while those in the LDA-based methods correspond to tokens in the documents. The use of word variables allows the detection and representation of patterns of word co-occurrences qualitatively using model structures as illustrated in Figure~\ref{fig.hltm}.
 When token variables are used, on the other hand, we cannot have a graphical structure among words because they are  values of the nodes (tokens) rather than nodes themselves. It is clear from   Figure~\ref{fig.hltm} that modeling structures among words is conducive to the discovery of meaningful topics and topic hierarchies.

Another important difference is in the definition and characterization of topics. Topics in the LDA-based methods are probabilistic distributions over a vocabulary. When presented to users, a topic is characterized using a few words with the highest probabilities. In contrast, topics in HLTA/HLTA-c are clusters of documents. For presentation to users, a topic is characterized using the words that not only occur with high probabilities in the topic but also occur with low probabilities outside the topic.

A third difference lies in the relationship between topics and documents. In the LDA-based methods, a document is a mixture of topics, and the probabilities of the topics within a document sum to 1. Because of this, the LDA models are sometimes called {\em mixed-membership models}. In HLTA/HLTA-c, a topic is a soft cluster of documents, and a document might belong to multiple topics with probability 1. In this sense, HLTMs can be said to be {\em multi-membership models}.

HLTA and HLTA-c produce hierarchies with word variables at the bottom and multiple levels of latent variables on top. It is related to hierarchical variable clustering. However, there are fundamental differences. One difference is that HLTA and HLTA-c define a distribution over documents while variable clustering does not.

HLTA and HLTA-c partition
document collections in multiple ways.  There is a vast literature on document
clustering. In particular, co-clustering~\cite{dhillon2001co} can identify document clusters where
each cluster is associated with a potentially different set of words. However,
document clustering and topic detection are generally considered two different fields with little overlap. This paper bridges the two fields by developing a full-fledged hierarchical topic detection method that partitions documents in multiple ways.

\begin{figure*}[t]
\centering
\subfigure[${M}_b$]{\includegraphics[width=4cm]{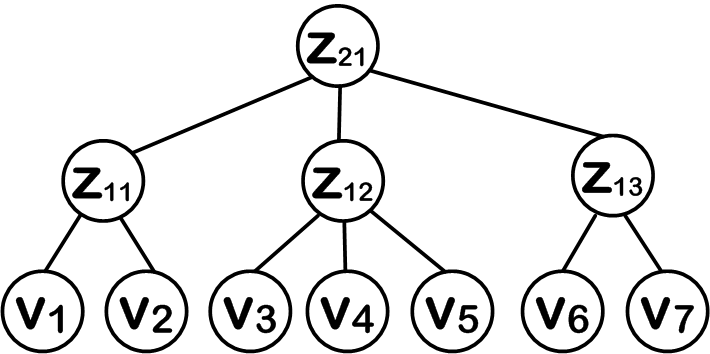}
\label{fig.hlta-b}}
\subfigure[${M}_c$]{\includegraphics[width=4cm]{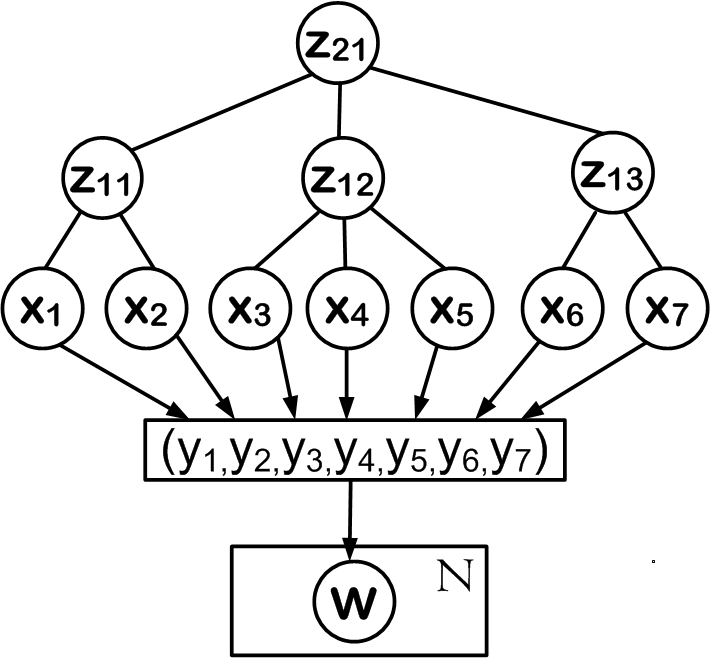}
\label{fig.docGen}}
\subfigure[${M}_a$]{\includegraphics[width=4cm]{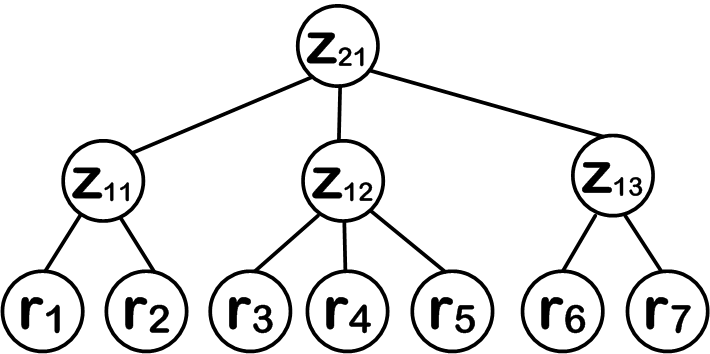}
\label{fig.hlta-a}}
\caption{\small The model  $M_b$  is learned by HLTA from binary data. Each $v_i$ is binary variable and corresponds to the $i$-th word in the vocabulary.  Its distribution given its parent is a Bernoulli distribution. The model ${M}_c$ is for modeling count data and defines a document generation process. Each unnormalized relative frequency variable $x_i$ takes its value from the interval $[0, 1]$ and its distribution given its parent is a truncated Normal distribution. The relative frequency variables $y_i$  are obtained by normalizing the $x_i$'s. They form  a multinomial distribution from which  words of a document are drawn.  The model ${M}_a$ is a auxiliary model used in parameter estimation and model evaluation. It is the same as the top part of ${M}_c$, except that the variables $r_i$ are observed and are not restricted to the interval $[0, 1]$. It shares the same parameters as ${M}_c$. }

\label{fig.comparison}
\end{figure*}

\section{Document Generation Process}
 Our HLTM-c model  for document generation  is illustrated in Figure \ref{fig.comparison} (b). It is based on an HLTM $M_b$ learned from binary data.

We regard $M_b$  as model that generates binary vectors over words as follows:
First, pick values for the latent variables using logic sampling
\cite{pear1988probabilistic}. Specifically, sample a value for the root $z_{21}$ from its marginal distribution $P(z_{21})$ \footnote{When there are multiple latent variables at the top level, arbitrarily pick one of them as the root.} and sample values for other latent variables $z_{11}, z_{12}, z_{13}$  from their conditional distributions given their respective parents. For example, the value for $z_{11}$ is sampled from the distribution $P(z_{11}|z_{21})$.  Then, sample a binary value for each of the  word variables $v_1, \ldots, v_7$ from the its conditional distribution given its parent. For example, the value for $v_{1}$ is sampled from the distribution $P(v_1|z_{11})$. Note that all distributions mentioned here are Bernoulli distributions.

In the HLTM-c model $M_c$, we generate values for the latent variables in the same way as in $M_b$.
  However, we do not sample binary values for word variables. Instead, we draw a real value $x_i$ for each word $v_i$. The value $x_i$ is restricted to lie in the interval $[0, 1]$ and it is meant to be, approximately, the relative frequency of the word in a document. Let $z$ be the parent of $x_i$.
We draw $x_i$ from a truncated normal distribution $$P(x_i|z) = {\cal TN}(\mu_{iz}, \sigma_{iz}^2, 0, 1)$$ \noindent with mean $\mu_{iz}$ and standard deviation $\sigma_{iz}$.
Note that, for a fixed $x_i$,  the notation $\mu_{iz}$ refers to two numerical values, one for $z=0$ and another for $z=1$.  The same can be said for $\sigma_{iz}$.

Different $x_i$'s are drawn independently and hence there is no guarantee that they sum to 1. We normalize them to get a multinomial distribution $(y_1,  \ldots, y_7)$. Finally, we draw words $W$ for a document from this multinomial distribution.

In general, suppose there is a collection $D$ of documents and there are $V$ words in the vocabulary. Assume an HLTM $M_b$ with binary word variables has been learned from the binary version of the data using the HLTA algorithm described in \cite{chen2017latent}.  To take word counts into consideration,
  we  turn $M_b$ into a document generation model $M_c$ by replacing the binary word variables $v_i$ with  real-valued variables $x_i$, and we assume a document of length $N$ is generated using  $M_c$ as follows:
 \begin{enumerate}
\item Draw values of the binary  latent variables via logic sampling.
\item For each $i \in \{1, \ldots, V\}$, draw $x_i$ from the conditional distribution $p(x_i|pa(x_i))$, a truncated normal distribution, of $x_i$ given its parent $pa(x_i)$.
\item For each $i \in \{1, \ldots, V\}$, set $y_i = {x_i}/{\sum_{i=1}^nx_i}$.
\item  For each $n \in \{1, \ldots, N\}$,  draw the $n$-th word  of the document from $Multi(y_1, \ldots, y_n)$.
\end{enumerate}

 In the generation process, Step 2 generates the {\em unnormalized relative frequencies (URF)} $\x=(x_1, \ldots, \x_V)$, while Step 3 obtains the (normalized) {\em relative frequencies} $\y = (y_1, \ldots, y_n)$. The first three steps  define a distribution over all possible the relative frequency vectors,  i.e., over the probability simplex $S=\{(y_1, \ldots, y_n)| y_i \geq 0, \sum_{i=1}^n y_i=1\}.$   We denote the distribution as $p(\y|{M}_c, \theta)$. The parameter vector $\theta$ includes the parameters for the distributions of all binary latent variables, and the means and standard deviations for the truncated normal distributions for the URF variables.

Let  $N_i$ be the number of times the $i$-th word from  the vocabulary  occurs in a document $d$. The count vector
 $\bc = (N_1, N_2, \ldots, N_V)$ can be used as a representation of $d$. The sum of those counts is the document length, i.e.,  $N= \sum_{i=1}^V N_i$. The conditional probability of $d$ given the relative frequencies $\y$ is:
\begin{eqnarray}
\label{eq.probdgivenasample}
P(d|\y) = \frac{N!}{N_1! \ldots N_V!}\prod_{i=1}^Vy_i^{N_i}.
\end{eqnarray}

The entire generation process defines a  distribution over documents.
The probability of a document $d$ is:
\begin{eqnarray}
\label{eq.probdgivenmodel}
P(d|{M}_c, \theta_c) = \int_{\y} P(d|\y) p(\y|{M}_c, \theta) d \y.
\end{eqnarray}

 The document generation process
given here is very different in flavor from the generation processes one typically sees in the LDA literature. Nonetheless, it is a well-defined generation process. It defines a distribution over  count-vector representations of documents.
  An LDA-based model, on the other hand,  defines a distribution over  bag-of-words representations of documents. Because the count-vector representation is equivalent to the bag-of-words representation, HLTM-c and  the LDA-based methods  define distributions over the same collection of objects and hence can be compared with each other.

\section{Parameter Estimation}
We now consider how to estimate the parameters $\theta$ of the mode $M_c$. The log likelihood function of
$\theta$ given a collection of documents ${D}$ is:
\begin{eqnarray}
\label{eq.probDocs}
\log P({D}|{M}_c, \theta) = \sum_{d \in {D}} \log P(d|{M}_c, \theta).
\end{eqnarray}
The objective of parameter estimation is to find the value of $\theta$ that maximizes the likelihood function. This task is difficult because of  the the use of truncated normal distributions and the normalization step   in the document generation process.

We propose an approximate method based on two ideas. First, notice that the model parameters $\theta$ influence the relative frequencies $\y$ and the word counts $\bc$ indirectly through the URF variables $\x$. Given $\x$, $\bc$ and $\y$ are independent of $\theta$. Our first idea is to obtain a point estimate of $\x$ from $\bc$ and regard $\x$ as observed variables afterwards.

It is well known that, given the word counts $\bc$, the maximum likelihood estimation (MLE) of $\y$ is $y_i=N_i/N$ for all $i$, i.e., the empirical relative word frequencies. Ignoring the normalization step, we also use the empirical relative word frequencies as a point estimation for URF variables $\x$, i.e., we assume $\x = (N_1/N, \ldots, N_V/N)$.

The second idea is to relax the restriction that $x_i$ must be from the interval $[0, 1]$, and to assume that $x_i$ is
sampled from a  normal distribution ${\cal N}(\mu_{iz}, \sigma_{iz}^2)$ instead of a truncated normal distribution ${\cal TN}(\mu_{iz}, \sigma_{iz}^2, 0, 1)$.

Those two considerations turn the problem of estimating $\theta$ in $M_c$ with data represented as count vectors into the problem of estimating $\theta$ in a related model, denoted as $M_a$,  with data represented as vectors of relative frequencies. As shown in Figure~\ref{fig.hlta-a},
 the auxiliary model $M_a$ is the same as the top part of ${M}_c$, except that the URF variables $x_i$'s are replaced with real-value variables $r_i$'s. The conditional distribution   of each $r_i$ given its parent $z$ is a normal distribution, i.e., $$p(r_i|z) ={\cal N}(\mu_{iz}, \sigma_{iz}^2).$$

 Use  $d_f$ to denote the vector of relative word frequencies in a document $d$, i.e., $d_f = (N_1/N, \ldots, N_V/N)$. Moreover, use   ${D}_f$ to denote the entire data set when represented as vectors of relative frequencies.  In the auxiliary model $M_a$, the log likelihood  of $\theta$ given $D_f$ is
\begin{eqnarray}
\label{eq.probDocs_a}
\log P({D}_f|{M}_a, \theta) =\sum_{d_f \in {D}_f} \log P(d_f|{M}_a, \theta).
\end{eqnarray}
Maximizing this likelihood function is relatively easy because $M_a$ is a tree model. It can be done  using the EM algorithm.

 We do not have any error bounds relating the target likelihood function $\log P({D}|{M}_c, \theta)$ and the approximate likelihood function
 $\log P({D}_f|{M}_a, \theta)$. However, there are strong reasons to believe that
 maximizing (\ref{eq.probDocs_a}) would result in high quality parameter estimation for the generative model $M_c$ due to the way the approximation is derived.
 Empirical results to be presented later show that the method does produce good enough parameter estimations for $M_c$ to  achieve substantially higher held-out likelihood than the LDA-based methods.

 Although $M_a$ is a tree model, EM can still be very time consuming
 when the sample size is large. In this case,  we use stepwise EM \cite{stepEM2000,stepEM2009}, which is the result of applying the idea of stochastic gradient descent to EM. It scales up much better than EM.

\section{Model Evaluation}
After obtaining an estimation $\theta^*$ of the parameters in the document generation model ${M}_c$, we need to evaluate it by calculating its log likelihood on a test set ${D}_t$:
\begin{eqnarray*}
\label{eq.likelihood_test}
\log P({D}_t|{M}_g, \theta^*)) =\sum_{d \in {D}_t} \log P(d|{M}_c, \theta^*).
\end{eqnarray*}

To calculate the probability $\log P(d|{M}_c, \theta^*)$ of a test document $d$, we need to approximately compute the integration in  (\ref{eq.probdgivenmodel}). Since the distribution of $P(\y|{M}_c, \theta^*)$ is defined through a generation process, it is straightforward  to obtain  samples of $\y$ by running the process multiple times.  Suppose $K$ samples $\y^{(1)}, \ldots, \y^{(K)}$ of $\y$ are obtained. We can estimate
$P(d|{M}_c, \theta^*)$  as follows:
\begin{eqnarray}
\label{eq.estProbdocgivenmodel}
P(d|{M}_c, \theta^*) \approx  \frac{1}{K} \sum_{k=1}^KP(d|\y^{(k)}).
\end{eqnarray}

Unfortunately, there is a well-known problem with this naive method~\cite{wallach2009evaluation}.
When the document $d$ is long, the integrand $p(d|\y)$ as a function of $\y$ is highly peaked around the MLE of $\y$ and is very small elsewhere in the probability simplex. Because the document $d$  is not taken into consideration
when drawing  samples of $\y$,  it is unlikely for the samples to hit the high value area. This can easily lead to underestimation and  high variance. Unless $K$ is extremely large, there could be large differences in the estimates one obtains at different runs.

A standard way to solve this problem is to use {\it importance sampling} \cite{owen2013monte}, and to utilize a proposal distribution that is related to $d$ and has its density concentrated in the region where the integrand function  $P(d|\y)$ is not close to zero.
We derive such a distribution using the auxiliary model $M_a$.

{\rowcolors{2}{white}{gray!20}
\begin{table*}[t]
\centering
\caption{Per-document held-out log likelihood scores. The sign ``-'' indicates non-termination after 96 hours.}
\setlength\tabcolsep{3pt}
  \begin{tabular}{l|ccccccc}
        & {\scriptsize{NIPS-1k}}& {\scriptsize{NIPS-5k}}& {\scriptsize{NIPS-10k}}&   {\scriptsize{News-1k}}&   {\scriptsize{News-5k}} & {\scriptsize{NYT}}&{\scriptsize{AI}} \\ \hline
    HLTA-c & {\bf -1,182}$\pm$2 &{\bf -2,658$\pm$1} &{\bf -3,249$\pm$2} & {\bf -183$\pm$1} & -{\bf 383$\pm$2}& {\bf-1,255$\pm$3} & {\bf-3,216$\pm$3}\\
    hLDA &-2,951$\pm$35 &-5,626$\pm$117& --- &---&--- & --- & --- \\
    nHDP& -3,273$\pm$6 &-7,169$\pm$11 &-8,318$\pm$18 & -262$\pm$1 & -565$\pm$3&-2,070$\pm$6 & -7,606$\pm$ 12 \\
    hPAM& -3,196$\pm$3 &-6,759$\pm$15 & -7,922 $\pm$ 12 &-255$\pm$2 & -556$\pm$4 &--- &---\\
\end{tabular}
\label{tbl:perll}
\end{table*}
}

Let $\z$ be the set of all latent variables in the auxiliary model ${M}_a$ on the level right above the $r_i$'s.
For each latent variable $z$ in $\z$, it is easy to compute the posterior distribution  $p(z|d_f, {M}_a, \theta^*)$ of $z$  given a document $d$ (represented as a vector of relative word frequencies $d_f$) because $M_a$ is a tree model. In fact, it can be done in linear time using message propagation. We define
\begin{eqnarray}
q(\z|d) =\prod_{z \in \z} p(z|d_f, {M}_a, \theta^*).
\label{eq.q}
\end{eqnarray}

Note that $\z$ (defined in $M_a$) is the same as  the set of all latent variables in the generative model $M_c$ on the level right above the $x_i$'s.
We  rewrite (\ref{eq.probdgivenmodel}) as follows for the test document $d$:
\begin{eqnarray*}
P(d|{M}_c, \theta^*) &=&\int \sum_{\z}P(d|\y)p(\y|\z) p(\z|{M}_c, \theta^*) d\y
\end{eqnarray*}

\noindent
Note that $p(\z|{M}_c, \theta^*) = p(\z|{M}_a, \theta^*)$. Inserting $\frac{q(\z|d)}{q(\z|d)}$ into the right hand side and rearranging terms, we get
\begin{eqnarray*}
P(d|{M}_c, \theta^*)   &=&  \int \sum_{\z}P(d|\y) \frac{p(\z|{M}_a, \theta^*)}{q(\z|d)} p(\y|\z)q(\z|d) d \y.
\end{eqnarray*}

\noindent
This expression implies that we can sample a sequence of pairs $(\y^{(1)}, \z^{(1)}), \ldots, (\y^{(K)}, \z^{(K)})$  from $p(\y|\z)q(\z|d)$  \footnote{For each pair, first sample $\z$ from $q(\z|d)$, and then sample $\y$ from  $p(\y|\z)$}, and estimate  $P(d|{M}_c, \theta^*)$ as follows:
\begin{eqnarray}
\label{eq.estProbdocgivenmodel2}
P(d|{M}, \theta^*) & \approx& \frac{1}{K}  \sum_{k=1}^KP(d|\y^{(k)}) \frac{p(\z^{(k)}|{M}_a, \theta^*)}{q(\z^{(k)})}.
\end{eqnarray}
 Note that $P(d|\y^{(k)})$ can be calculated using (\ref{eq.probdgivenasample}) and the term $p(\z^{(k)}|{M}_a, \theta^*)$  is obtained using message propagation in $M_a$. As mentioned earlier,  the term $q(\z^{(k)})$ can also be easily computed in $M_a$.

In comparison with (\ref{eq.estProbdocgivenmodel}), the use of (\ref{eq.estProbdocgivenmodel2}) improves estimation accuracy and reduces the variance because the sample points
$(\y^{(k)}, \z^{(k)})$ are generated by taking the test document $d$ into consideration.  Hence, the samples are more likely to hit the area where the integrand function $P(d|\y)$ has high values. In the experiments, we set $K=300$.

\section{Empirical Results}

We have now finished describing the HLTA-c algorithm.  Starting with a collection of documents, it first learns a model by running HLTA on the binary version of the data. Then it turns the model into a document generation model and estimates the parameters from the count version of the data. The final model defines a probability distributions over count-vector representations of documents and gives a hierarchy of topics.

In this section, we present empirical results to compare HLTA-c with common LDA-based methods for hierarchical topic detection, including hPAM~\cite{mimno2007mixtures}, hLDA~\cite{blei10nested} and nHDP~\cite{paisley2015nested}. The comparisons are in terms of both  model quality  and the quality  of topics and topic hierarchies. Model quality is measured using held-out likelihood on test data. The quality of topics is assessed using
  topic coherence~\cite{mimno2011optimizing} and topic compactness~\cite{chen2016sparse}.  Example branches of the topic hierarchies obtained by  nHDP and HLTA-c  are also included for qualitative comparisons.

{\rowcolors{2}{white}{gray!20}
\begin{table*}[t]
\centering
\caption{Average coherence scores.}
\setlength\tabcolsep{1pt}
  \begin{tabular}{l|ccccccc}
        & {\scriptsize{NIPS-1k}}& {\scriptsize{NIPS-5k}}& {\scriptsize{NIPS-10k}}&   {\scriptsize{News-1k}}&   {\scriptsize{News-5k}} & {\scriptsize{NYT}} & {\scriptsize{AI}} \\ \hline
    HLTA-c& {\bf -6.46$\pm$0.01}&{\bf -8.20$\pm$0.02}&{\bf -8.93$\pm$0.04}&-12.50$\pm$0.08& {\bf -13.43$\pm$0.15} & {\bf -12.70 $\pm$ 0.18}  & {\bf -16.18  $\pm$ 0.15} \\
    hLDA & -7.46$\pm$0.31 &-9.03$\pm$0.16 &--- & ---&---& --- &--- \\
    nHDP & -7.66$\pm$0.23 & -9.70$\pm$0.19  &-10.89$\pm$0.38 & -13.51$\pm$0.08& -13.93$\pm$0.21& -12.90$\pm$0.16& -18.66$\pm$0.21\\
    hPAM &  -6.86$\pm$0.08 & -8.89$\pm$0.04 & -9.74$\pm$0.04 & {\bf -11.74$\pm$0.14} & -14.06 $\pm$0.09& ---& ---\\

    \end{tabular}
\label{tbl:coherence}
\end{table*}
}

{\rowcolors{2}{white}{gray!20}
\begin{table*}[t]
\centering
\caption{Average compactness scores.}
\setlength\tabcolsep{1.1pt}
  \hspace{-0.5cm}
  \begin{tabular}{l|ccccccc}
        & {\scriptsize{NIPS-1k}}& {\scriptsize{NIPS-5k}}& {\scriptsize{NIPS-10k}}&   {\scriptsize{News-1k}}&   {\scriptsize{News-5k}} & {\scriptsize{NYT}} & {\scriptsize{AI}}\\ \hline
    HLTA-c& {\bf 0.228$\pm$0.001}&{\bf 0.255$\pm$0.001}&{\bf 0.243$\pm$0.001}& {\bf 0.219$\pm$0.001}& {\bf 0.226$\pm$0.001 }& {\bf 0.288$\pm$0.009} &{\bf  0.229$\pm$0.001}\\
   hLDA & 0.163$\pm$0.003 &0.153$\pm$0.001 &--- & ---&---& ---& ---   \\
      nHDP & 0.164$\pm$0.005 &0.147$\pm$0.006 &0.138$\pm$0.002 & 0.150$\pm$0.003 & 0.148$\pm$0.004 & 0.250$\pm$0.003& 0.144$\pm$0.001\\
    hPAM &0.211$\pm$0.003 &0.167$\pm$0.001 &0.141$\pm$0.002 &0.210$\pm$0.006 &0.178$\pm$0.002& ---& --- \\
    \end{tabular}
\label{tbl:Compactness}
\vspace{-3mm}
\end{table*}
}

\begin{figure*}[t]
\centering
\includegraphics[width=16cm]{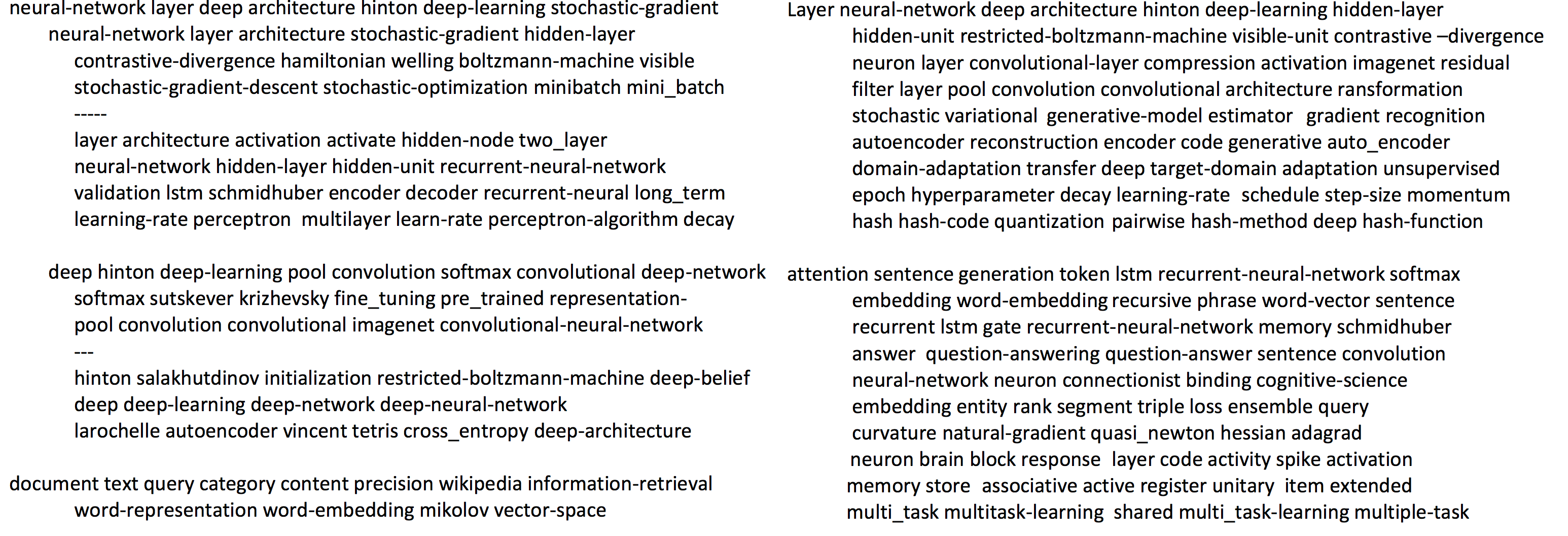}
\caption{\small Selected branches of the topic trees produced by HLTA-c (left) and nHDP (right) on the AI dataset. }
\label{fig.topics}
\end{figure*}

\subsection{Datasets and Settings}
We  used four datasets in our experiments. The first one is the  NIPS dataset,  which consists of 1,955 articles published at the NIPS conference between 1988 and 1999\footnote{\scriptsize {\tt http://www.cs.nyu.edu/{$\sim$}roweis/data.html}}. The second one is the 20 Newsgroup dataset\footnote{\scriptsize \tt http://qwone.com/$\sim$jason/20Newsgroups/}, which consists of 19,940 newsgroup posts.  The third one is the New York Times (NYT) dataset\footnote{\scriptsize {\tt http://archive.ics.uci.edu/ml/datasets/Bag+of+Words}}, which consists of 300,000 articles published on New York Times between 1987 and 2007. The last one is a dataset that includes all 24,307 papers published at seven AI conferences and three AI journals  between 2000 to 2017. We refer to it as the AI dataset.

 To have some variabilities on the vocabulary size, we created different versions of
 the NIPS and the Newgroup datasets by choosing vocabularies with different sizes using  average TF-IDF.
  The NIPS dataset has three versions with vocabulary sizes of 1,000, 5,000 and 10,000 respectively, and the Newsgroup dataset has two versions   with vocabulary sizes of 5,000 and 10,000. The NYT and AI datasets each have only one version with vocabulary size  10,000.

Implementations of HLTA and the LDA baselines were obtained from their authors
\footnote{\tiny \tt github.com/kmpoon/hlta; github.com/blei-lab/hlda; www.columbia.edu/$\sim$jwp2128/code/nHDP.zip; www.arbylon.net/projects/knowceans-lda-cgen/Hpam2pGibbsSampler.java.}.
HLTA-c was implemented on top of HLTA. The implementation will be released
along with the publication of this paper.

HLTA-c determined the height of topic hierarchy and the number of nodes at
each level by running the HLTA at its default parameter settings on the
binary version of a dataset. We tuned the parameters of the LDA-based
baselines in such a way that they would yield roughly the same total number
of topics as HLTA-c.
For example, HLTA produced 1133 topics on the AI dataset. To ensure that
nHDP would produce a similar number of topics, we used $20$ nodes for the
top level, $20 \times 6$ nodes for the the second level, and $20 \times 6 \times 8$ for the third level, leading to a total number of 1100 topics.
%
%
The other parameters of the baselines that do not affect the number of topics were left at their default values.

HLTA-c needs to call stepwise EM in the parameter estimation step. Stepwise EM has a parameter called stepwise $\eta_t$, which we set as $\eta_t = (t+2)^{-0.75}$ as is usually done in the literature.
All experiments were conducted on the same desktop computer.

\subsection{Model Quality}

We randomly divided each dataset into a training set with $80\%$ of the data, and a test set with $20\%$ of the data.
The algorithms were run on the training set, and the resulting models were evaluated on the test set. The per-document log likelihood scores are reported in  Table~\ref{tbl:perll}.

We see that the held-out likelihood scores for HLTA-c are drastically higher than those for all the baseline methods. On the NYT and AI datasets, for instance, the models produced by HLTA-c have scores of -1,255 and -3,216 respectively, while the models by nHDP have scores of -2,070 and -7,606. The results imply that
 the models obtained by HLTA-c can predict unseen data much better than those by the other methods.

HLTA-c not only achieved much higher held-out likelihood scores than the baselines, but also did so with much fewer parameters. In an HLTM-c, each URF variable $x_i$  has 4 parameters. Suppose there are $V$ words in the vocabulary. The total number of parameters associated with those variables is $4V$. The number of latent variables is upper bounded by $V$ and each of them has no more than 2 parameters. So, the total number of parameters is upper bounded by $6V$.  In an LDA-based model with $\beta$ topics, on the other hand, one needs $\beta (V-1)$ parameters just to describe the topics.  In practice, $\beta$ is much larger than 6.
In addition, one also needs to describe the topic proportions for each document. The number of parameters needed here increases linearly with the number of documents, which can be very large.

\subsection{Topic Extraction from HLTM-c}

Suppose an HLTM-c has been learned from a collection of documents.
Each latent variable $z$ in the model partitions the collection into two soft clusters.
To interpret the clusters, we consider all the URF variables in the subtree rooted at $z$ and examine their distributions in the two clusters. Let $x_i$ be such a variable, and let $\mu_i$ and $\nu_i$ be the means of $x_i$ in the two clusters. We sort the variables in descending order of the difference $|\mu_i - \nu_i|$, and pick a number of top variables to characterize the differences between the two clusters.

The following table shows the information about a latent variable in the HLTM-c learned from the AI dataset: \footnote{At the preprocessing stage, frequent n-grams in the AI dataset were combined into tokens.}

\vspace{0.1cm}

{\scriptsize
\begin{center}
\begin{tabular}{|l|c|c|} \hline
                & cluster 1 (0.93)     & cluster 2 (0.07) \\ \hline
\word{neural-network}   & $0.00029 \pm 1.5E-3$             & $0.00455 \pm 6.9E-3$ \\
\word{layer}   & $0.00069  \pm 3.5E-3$             & $0.00471 \pm 9.5E-3$ \\
\word{deep}   & $0.00012 \pm 9.7E-4$             & $0.00301 \pm 5.4E-3$ \\
\word{architecture}   & $0.00053 \pm 2.1E-3$             & $0.00324 \pm 5.5E-3$ \\
\word{hinton}   & $0.00011 \pm 6.3E-4$             & $0.00187 \pm 2.4E-3$ \\
\word{deep-learning}   & $0.00004 \pm 5.5E-4$             & $0.00168 \pm 3.3E-3$ \\
\ldots  & \ldots              & \ldots \\
\hline
\end{tabular}
\end{center}
}

\vspace{0.1cm}

It partitions the documents into two clusters that consist of 93\% an 7\%  of the documents respectively. In the second cluster, the URF variables have relatively high means, and hence the cluster is interpreted as a topic. The topic is labeled with the top words,
i.e., ``{\tt neural-network} {\tt layer} {\tt deep} {\tt architecture}
 {\tt hinton} {\tt deep-learning}". Clearly, the topic is about  {\tt neural network} and {\tt deep learning}. In the first cluster, the URF variables have relatively low means, and hence the cluster is interpret as background, consisting of documents not in the topic just mentioned.

\subsection{Topic Quality}

Both HLTA-c and the LDA-based methods characterize topics using lists of words when presenting them to users. Direct comparisons are therefore possible. We measure the quality of a topic using two metrics. The first one is the
 \emph{topic coherence score}~\cite{mimno2011optimizing}.
The intuition behind this metric is that words in a good topic should tend to co-occur in the documents.
 Suppose a topic $t$ is characterized by a list $\{w_1, w_2, \ldots, w_M\}$ of  $M$ words. The coherence score of $t$ is given by:
  \begin{eqnarray*}
  {\small
  {\tt coherence}(t) = \sum\limits_{i=2}^{M}\sum\limits_{j=1}^{i-1}\log \frac{D(w_i,w_j)+1}{D(w_j)},}
  \end{eqnarray*}

\noindent where $D(w_i)$ is the number of documents containing the word $w_i$, and $D(w_i, w_j)$ is the number of documents containing both $w_i$ and $w_j$. Higher coherence score means better topic quality.

The second metric is the \emph{topic compactness score}~\cite{chen2016sparse}. It is calculated on the basis of the word2vec model that
was trained on a part of the Google News dataset\footnote{\scriptsize  \tt https://code.google.com/archive/p/word2vec/} \cite{mikolov2013efficient,DBLP:conf/nips/MikolovSCCD13}. The word2vec model maps each word into a vector that captures the semantic meaning of the word.
The intuition behind the compactness score   is that words in a good topic should be closely related semantically.
The compactness score of a topic  $t$ is given by:
\begin{eqnarray*}
{\small
{\tt compactness}(t) = \frac{2}{M(M-1)}\sum\limits_{i=2}^{M}\sum\limits_{j=1}^{i-1} S(w_i,w_j),}
\end{eqnarray*}

\noindent where $S(w_i, w_j)$ is the cosine similarity between the vector representations of the words $w_i$ and $w_j$. Words that do not occur in the  \emph{word2vec} model  were simply skipped. Higher compactness score means better topic quality.

Both of the scores decrease with the length $M$ of the word list. Some of the topics produced by HLTA-c consist of only 4 words. Hence, we set $M=4$. Using a higher value for $M$ would put the LDA-based methods at a disadvantage.

The average coherence and compactness scores are shown in
 Tables~\ref{tbl:coherence} and \ref{tbl:Compactness}.
 We see that HLTA-c achieved the highest compactness score in all cases. It also achieved the highest coherence score in all cases except for News-1k. The differences between the scores for HLTA-c and those for the LDA-based methods are often large. On the AI dataset, for instance, HLTA-c achieved a coherence score of -16.18 and a compactness score  of 0.299. The corresponding scores for nHDP are -18.66 and 0.144 respectively.

\subsection{Selected Branches of Topic Hierarchies}

Figure \ref{fig.topics} shows branches of the topic trees produced by HLTA-c and nHDP on the AI dataset that are related to neural networks and deep learning. In the HLTA-c topic tree, there is a topic on {\tt neural network} and {\tt deep learning},
which has a subtopic on {\tt neural network} and another subtopic on {\tt deep learning}. The topic  {\tt neural network} in turn has subtopics on network {\tt architecture} and training algorithms ({\tt contrastive divergence} and {\tt stochastic gradient descent}). The topic on {\tt deep learning} has subtopics on {\tt convolutional neural network}, {\tt restricted Boltzmann machine}, {\tt deep neural network}, and {\tt autoencoder}. The names of several prominent  deep learning authors appear in the topic descriptions.  The topics {\tt recurrent neural network} and {\tt lstm} are placed in the first group instead of the second, which indicates that they co-occur more often with the {\tt neural network} topics than the {\tt deep learning} topics. In fact, in RNN papers, one comes across phrases such as {\tt neural network}, {\tt layer} and {\tt hidden layer} more often than phrases such as {\tt deep}, {\tt pool} and {\tt convolution}. A similar  statement can be made about  {\tt word embedding}.

The topics obtained by nHDP are also meaningful. However, the relationships among them are not as meaningful as those among the HLTA-c topics. In particular, the last three topics clearly do not fit well with the other topics in the group.

\section{Concluding Remarks}
HLTA is a recently proposed method for hierarchical topic detection.
It can only deal with binary data. In this paper we extend HLTA so as to take word counts into consideration. It is achieved by proposing a document generation process based on the model structure learned by HLTA. The extended method is called HLTA-c. In comparison with LDA-based methods, HLTA-c achieves far better held-out likelihood with much fewer parameters. It also produces significantly better topics and topic hierarchies. It is the new state-of-the-art for hierarchical topic detection.

In recent years, machine learning research mostly focuses on parameter learning. Relative little attention is placed on model structure learning. The work on HLTA and this paper serve to illustrate a strategy where one uses a simpler form of data for model structure learning and the full data for parameter learning. Such a strategy can lead to superior performances when compared with the practice where ones relies on manually constructed model structures.

\newpage

\bibliographystyle{aaai}
\bibliography{ref}

\end{document}